\documentclass[runningheads,a4paper]{llncs}

\usepackage{amssymb}
\usepackage{graphicx}
\usepackage[utf8]{inputenc}
\usepackage{url}
\usepackage{tabularx}
\usepackage{epstopdf}
\usepackage{graphicx}
\usepackage{caption}
\usepackage{amsmath}
\usepackage{subcaption}
\usepackage{subfig} 

\urldef{\mailsa}\path|{pplonski, zaremba}@ire.pw.edu.pl|    
\newcommand{\keywords}[1]{\par\addvspace\baselineskip
\noindent\keywordname\enspace\ignorespaces#1}

\begin{document}

\mainmatter 
\title{Visualizing Random Forest with \\ Self-Organising Map} 
\titlerunning{Visualizing Random Forest with Self-Organising Map.}
\author{Piotr Płoński \and Krzysztof Zaremba}
\authorrunning{P. Płoński and K. Zaremba}
\institute{Institute of Radioelectronics, Warsaw University of Technology,\\ Nowowiejska 15/19,00-665 Warsaw, Poland,\\
\mailsa}
\maketitle
\footnotetext{The final publication is available at  \url{http://link.springer.com/chapter/10.1007/978-3-319-07176-3_6}}

\begin{abstract}
Random Forest (RF) is a powerful ensemble method for classification and regression tasks. It consists of decision trees set. Although, a single tree is well interpretable for human, the ensemble of trees is a black-box model. The popular technique to look inside the RF model is to visualize a RF proximity matrix obtained on data samples with Multidimensional Scaling (MDS) method. Herein, we present a novel method based on Self-Organising Maps (SOM) for revealing intrinsic relationships in data that lay inside the RF used for classification tasks. We propose an algorithm to learn the SOM with the proximity matrix obtained from the RF. The visualization of RF proximity matrix with MDS and SOM is compared. What is more, the SOM learned with the RF proximity matrix has better classification accuracy in comparison to SOM learned with Euclidean distance. Presented approach enables better understanding of the RF and additionally improves accuracy of the SOM.

\end{abstract}

\section{Introduction}

Nowadays, there is a need for efficient data mining techniques. The human readability of the model is an important factor of a good data mining algorithm. Among various data mining methods very popular are decision trees \cite{quinlan}, \cite{marcin}. Although, they have an easy interpretable model, a single tree does not always obtain the highest accuracy. To overcome this problem, various ensemble methods were proposed. Among them, the popular is Random Forest (RF) proposed by Leo Breiman \cite{breiman}. The RF builds a set of trees using bagging and random subspace methods. The final output is a mode of responses from all individual trees. The RF can be used for classification and regression tasks. Despite the high accuracy of the RF, the human readability of the model is lost. There exist some methods to look inside RF black-box, like: examining variable importance \cite{var_sel}, parallel cooridinate plots by variable \cite{breiman} or visualizing the RF proximity distance matrix with Multidimensional Scaling \cite{liaw}. Herein, we propose a novel method for visualizing the RF proximity matrix based on Self-Organising Maps (SOM) \cite{kohonen}. The SOM is an artificial neural network model that maps high-dimensional input data space onto usually two-dimensional lattice of neurons in an unsupervised way. Although, the SOM is an originally unsupervised algorithm there exist supervised extensions \cite{lasso}, \cite{mh_som}, \cite{sto}, \cite{fuzzy_SOM}. The SOM has been proved as an efficient data mining tool in many real life applications \cite{dominik1}, \cite{dominik2}, \cite{julek}, \cite{julek2}. In this paper, we focus on using the labeled SOM model for mapping the RF used in classification tasks. The RF proximity matrix will be used for the SOM learning. It was shown that using more sophisticated distance metric than Euclidean can improve the accuracy of the SOM \cite{lasso_dml}. The RF proximity matrix was used earlier for improving clustering accuracy. Horvath et al. \cite{horvath} presented method for building clusters from the RF learned with unlabeled data and successfully used it for tumor detection \cite{horvath2}. Moosmann et al. \cite{moosmann} used the RF for efficient segmentation of images, where leaves were assigned to distinct image regions rather than to specific class. Gray et al. \cite{gray} used the RF proximity matrix and the MDS for classification of medical images of different types of dementia.
	The paper is organized as follow: firstly, we describe the SOM and the RF algorithms; secondly, proposed approach for learning the SOM with the RF proximity matrix is presented; then, the MDS vs  the SOM visualization and the accuracy of the SOM learned with Euclidean metric and the RF proximity matrix are compared. 

\section{Methods}

Let's denote data set as $D = \{ ( \vec{x_i}, c_i)\}$, where $\vec{x_i}$ is an attribute vector, $\vec{x} \in \mathcal{R}^M$, $M$ is attribute vector length and $c_i$ is a discrete class number of \textit{i}-th sample, $i=[1,2, ... , N]$ and $c=[1,2, ..., C]$. 

\subsection{Self Organising Maps} \label{som}

In this paper, we used the SOM as a two-dimensional grid of neurons. Each neuron is represented by a weights vector $W_{pq}$, where $(p,q)$ are indices of the neuron in the grid. It is important to notice, that neuron's weights vector has the same length as sample's attribute vector in the data set. The neuron's weights directly corresponds to attributes in the data set. In the training phase, for each sample we search for a neuron which is the closest to the \textit{i}-th sample. In the original SOM algorithm the distance is computed with squared Euclidean distance by following equation:
\begin{equation}\label{distance}
Dist_{train}(D_i, W_{pq}) = (\vec{x_i} - W_{pq})^T(\vec{x_i} - W_{pq}).
\end{equation}
The neuron $(p,q)$ with the smallest distance to \textit{i}-th sample is so-called the Best Matching Unit (BMU), and we note its indicies as $(r, v)$. Once the BMU is found, the weights update step is executed. The weights of each neuron are updated with formula:
\begin{equation}\label{learning}
W_{pq}(t+1) = W_{pq}(t) + \eta h_{pq}^{(i)} (\vec{x_i} - W_{pq}(t)),
\end{equation}
where $t$ is an iteration number and $\eta$ is a learning coefficient and $h_{pq}^{(i)}$ is a neighbourhood function. We assume that one iteration is a presentation of one training sample, whereas a presentation of all training samples is one learning epoch. The learning coefficient $\eta$ is decreased between consecutive epochs to improve network's ability to remember patterns. It is described by:
\begin{equation}
\eta = \eta_{0} exp(- e \lambda_{\eta}),
\end{equation}
where $\eta_{0}$ is the initial step size, $e$ is the current epoch number and $\lambda_{\eta}$ is responsible for regulating the speed of the decrease. The neighbourhood function controls changing of  weights with respect to the distance to the BMU in the grid. It is noted as:
\begin{equation}
h_{pq}^{(i)} = exp(-\alpha((r - p)^2 + (v - q)^2)),
\end{equation}
where $\alpha$ describes the neighbourhood function width. This parameter is increasing during learning $\alpha=\alpha_0 exp(-(e_{stop}-e)\lambda_{\alpha})$ - it assures that neighbourhood becomes narrower during the training. The network is trained till chosen number of learning procedure epochs $e_{stop}$ is exceeded. 

In the described algorithm the class label information is not used. The simplest approach of using the SOM as a classifier is to label the neurons after the unsupervised training. For each neuron we remember the overal sum  of neighbourhood values $h_{pq}^{(i)}$ from each class over all samples. The label of major class is assigned to the neuron. In the testing phase, the input sample's class is designated based on the class of the found BMU. 

\subsection{Random Forest}

In the RF algorithm a set of single trees is built. The process of constructing one tree can be described in the following steps:
\begin{enumerate}
\item Draw a bootstrap data set $D'$ by choosing $n$ times with replacement from all $N$ training samples. 
\item Determine a decision at node using only $m$ attributes, where $m$ is smaller than $M$. The split is selected based on maximal Information Gain. \label{dec}
\item Move the data through the node with respect to decision from the step \ref{dec}. \label{mov}
\item Repeat steps \ref{dec}, \ref{mov} till full tree is grown. 
\end{enumerate}
At the end of tree constructing, the class label is assigned to each leaf based on a class of samples in it. In the testing phase, a new sample is pushed down through all trees. From each tree a class label is remembered based on class of the reached leaf. The final response is the mode of votes from all trees. The proximity matrix $Prox$, with size $N$x$N$, can be easily obtained by putting all samples down the all trees. If two samples $i$ and $j$ are in the same terminal node in the tree, their proximity is increased by one $Prox(i,j) = Prox(i,j)+1$. After the presentation of all samples  the proximities are divided by the number of trees in the RF. The greater proximity value is, the more similar samples are. The dissimilarity measure can be formulated as $Dis(i,j) = 1 - Prox(i,j)$.

\subsection{Self Organising Maps learned with Random Forest (RF-SOM)}

In the proposed approach we assume that the RF is already learned. The learning of the network in one epoch can be summarized in the following steps:
\begin{enumerate}
\item Build a data set $H$ as a union of all network's weights $W$ and attribute vector $\vec{x}_j$ of $j$-th sample, $H = W \cup \vec{x}_j$. The matrix $W$ size is $L$x$M$, where the $L$ is a total number of neurons in the network. In this matrix each row contains weights from one neuron, the mapping from neurons's 2D grid to matrix $W$ is assumed. \label{fst}
\item For set $H$ compute dissimilarity matrix $Dis_H$ using the RF. The $Dis_H$ size is $(L+1)$x$(L+1)$.
\item Find the smallest distance to the neurons in dissimilarity matrix $Dis_H$, in distances corresponding to $j$-th sample:
\begin{equation}
v = \arg\min_h Dis_H (j,h), h \neq j,
\end{equation} 
where $v$ is an index of BMU in the matrix $W$, which can be mapped into $r,v$ indices in the network 2D grid. \label{bmu}
\item Update the network weigths with formula \ref{learning}. \label{lst}
\item Repeat steps \ref{fst}-\ref{lst} for all samples in the training set.
\end{enumerate}
After the end of the SOM learning, it is labeled as described in Section \ref{som}. In the testing phase, for input sample the BMU search is performed by taking the steps \ref{fst}-\ref{bmu}. The output class label is the same as the BMU class. We will denote the proposed method as RF-SOM. The computational complexity of using the Euclidean distance in the SOM is $\mathcal{O}(N * L)$, because we need to compute for each sample, from $N$ samples, a distance between sample and $L$ neurons. Whereas, the using of the RF proximity matrix in the proposed RF-SOM has complexity $\mathcal{O}(N*L*T*log_2(tree_{size}))$\footnote{We omit cost of constructing the RF in the complexity assessment.}, where $tree_{size}$ is a number of nodes in the tree. In the RF-SOM for each sample we propagate $L+1$ input vectors through $T$ trees in the RF, and passage through a tree has complexity $\mathcal{O}(log_2(tree_{size}))$. The complexity of distance computation using the RF proximity matrix is worse than using Euclidean distance. Although, it is beneficial when compared to the memory complexity of the MDS used for RF proximity visualization, which uses $\mathcal{O}(N^2)$ memory. The RF-SOM requires only $\mathcal{O}(L^2)$. This discards the MDS as a method of the RF proximity matrix visualization for large data sets. 

\section{Results}

We used 6 real data sets to examine properties of the RF-SOM. There were used data sets: 'Glass', 'Wine', 'Iris', 'Sonar', 'Ionosphere', 'Pima' from the 'UCI Machine Learning Repository' \cite{uci}. In the Table \ref{sets} are presented data sets properties. In all experiments we used following parameters values for each SOM type: $e_{stop} = 200$, $\eta_0 = 0.1$, $\lambda_{\eta}=0.0345$, $\alpha_0=0.1$, $\lambda_{\alpha}=0.008$. We will denote a network learned with Eculidean distance as SOM. The network sizes for the SOM and the RF-SOM for each data set are equal, they are presented in the Table \ref{sets}. The network sizes were chosen arbitrarily because selecting optimal network size is not in the scope of this paper. In all cases the SOM and the RF-SOM starts learning from the same initial weights values. The RF was constructed with 100 trees and $m=\sqrt{M}$ for all data sets.

\begin{table}
\begin{center}
\begin{tabular}{c|  >{\centering}m{1.7cm} | >{\centering}m{1.7cm} | >{\centering}m{1.7cm} | c |}
\cline{2-5}
 & Samples & Attributes & Classes & Network size \\ \hline 
\multicolumn{1}{|c|}{Glass} 		& 214  &  9  & 6 & 7x7 \\ \hline
\multicolumn{1}{|c|}{Wine}      	& 178  & 13  & 3 & 4x4 \\ \hline
\multicolumn{1}{|c|}{Iris}      	& 150  &  4  & 3 & 5x5 \\ \hline
\multicolumn{1}{|c|}{Sonar}       	& 208  & 60  & 2 & 8x8 \\ \hline
\multicolumn{1}{|c|}{Ionosphere} 	& 351  & 34  & 2 & 8x8 \\ \hline
\multicolumn{1}{|c|}{Pima}       	& 768  &  8  & 2 & 7x7  \\ \hline
\end{tabular}
\end{center}
\caption{The description of data sets used in experiments and network size used for each data set.}\label{sets}
\end{table}

\begin{figure}
\captionsetup[subfigure]{aboveskip=-13pt,belowskip=-3pt}
\centering
\begin{subfigure}[b]{0.49\textwidth}
	\caption{SOM} \label{som_pi}
	\includegraphics[trim=1.5cm 0cm 1.5cm 1.1cm, clip=true, width=\textwidth]{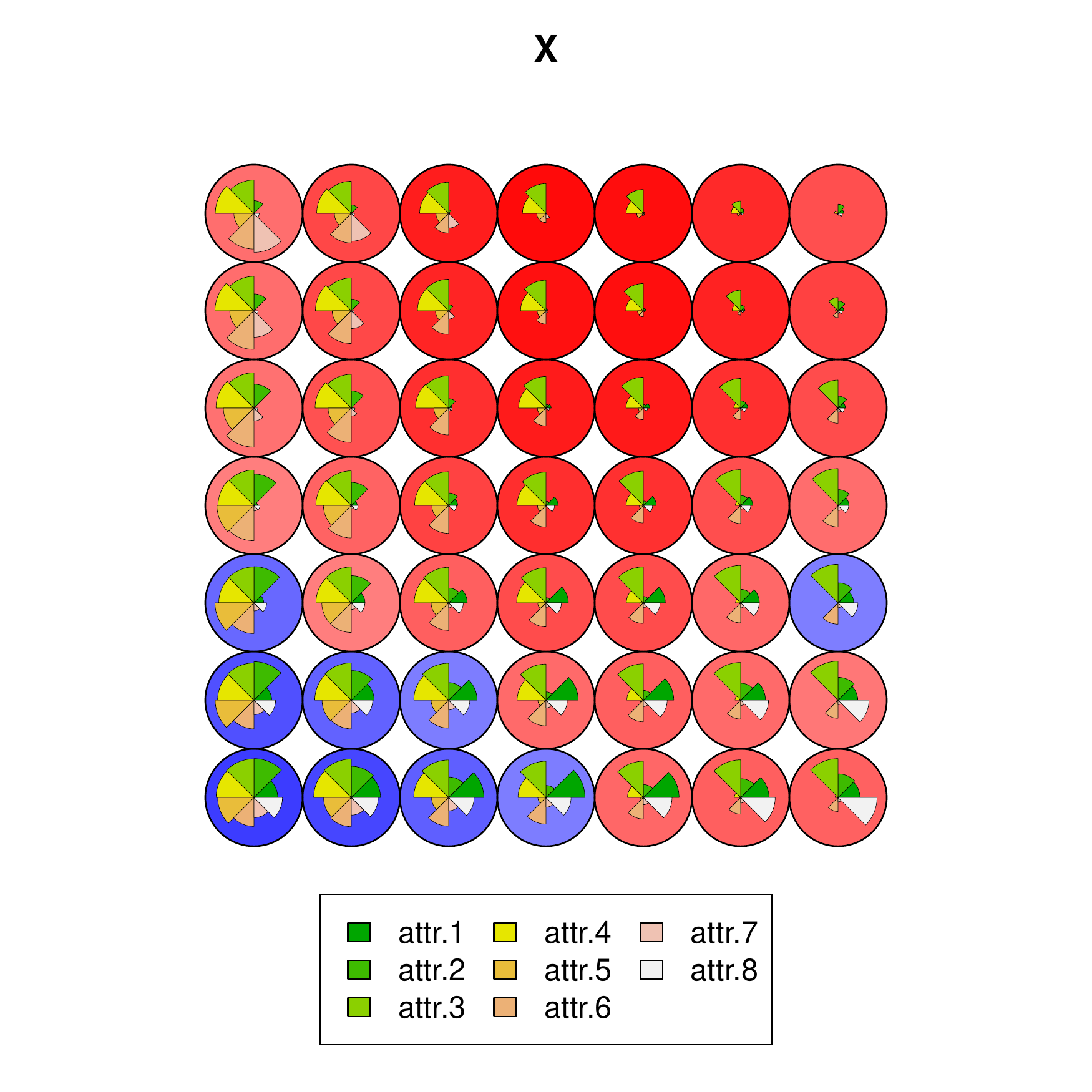} 		
\end{subfigure}
\begin{subfigure}[b]{0.49\textwidth}
	\caption{RF-SOM} \label{rfsom_pi}
	\includegraphics[trim=1.5cm 0cm 1.5cm 1.1cm, clip=true, width=\textwidth]{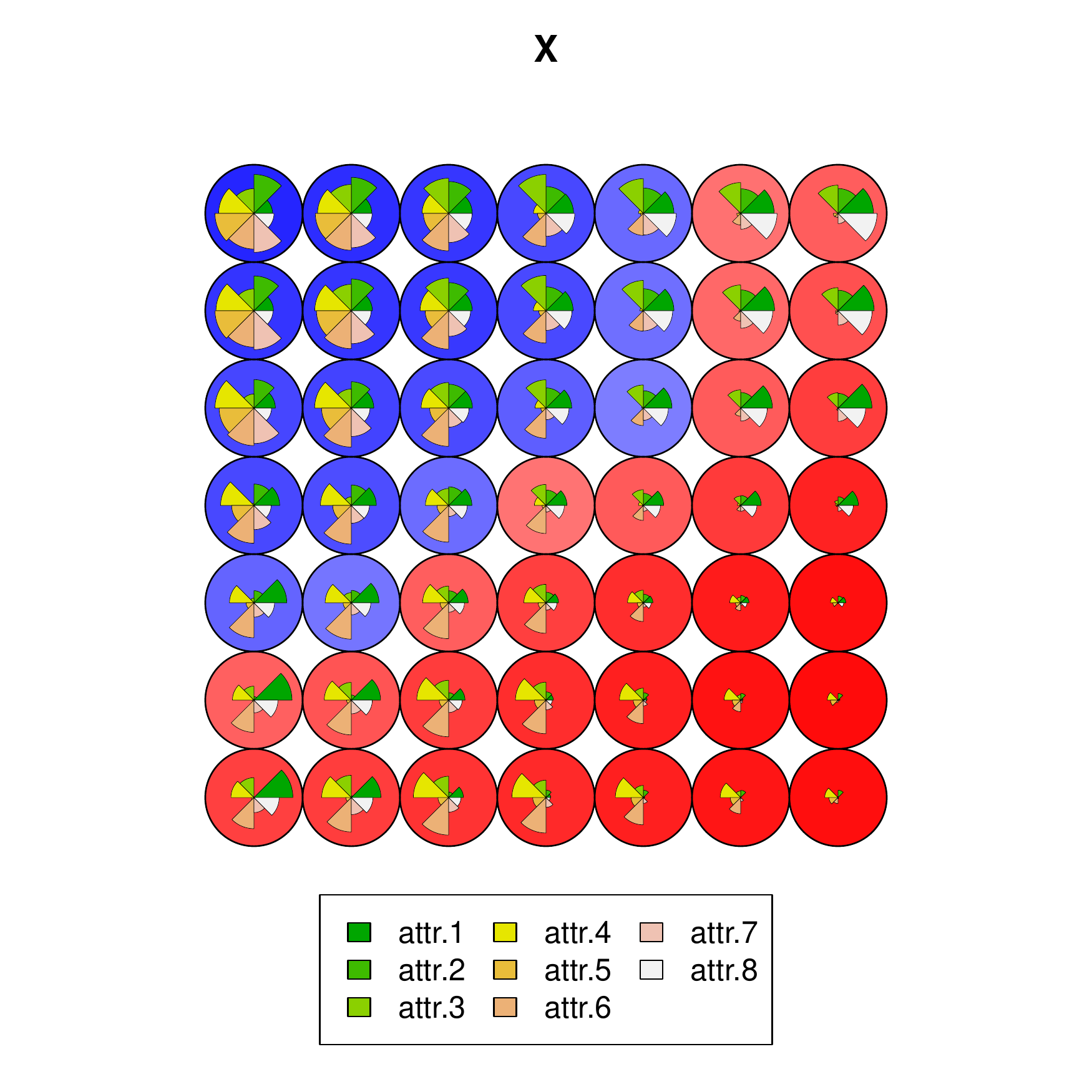} 
\end{subfigure}
\begin{subfigure}[b]{0.47\textwidth}
	\caption{MDS} \label{mds_pi}
	\includegraphics[clip=true, width=\textwidth]{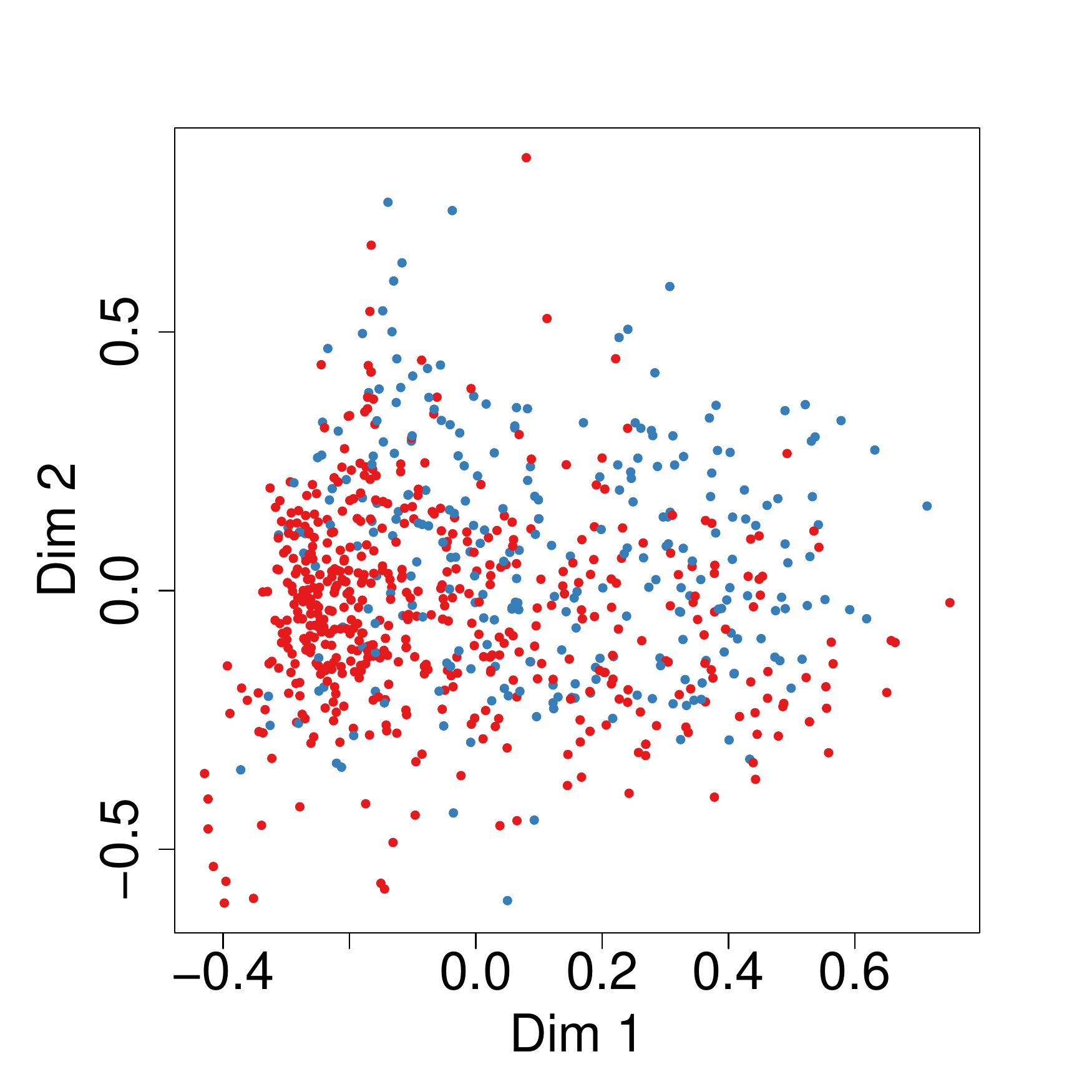}		
\end{subfigure}
\begin{subfigure}[b]{0.47\textwidth}
	\caption{RF-MDS} \label{rfmds_pi}
	\includegraphics[clip=true, width=\textwidth]{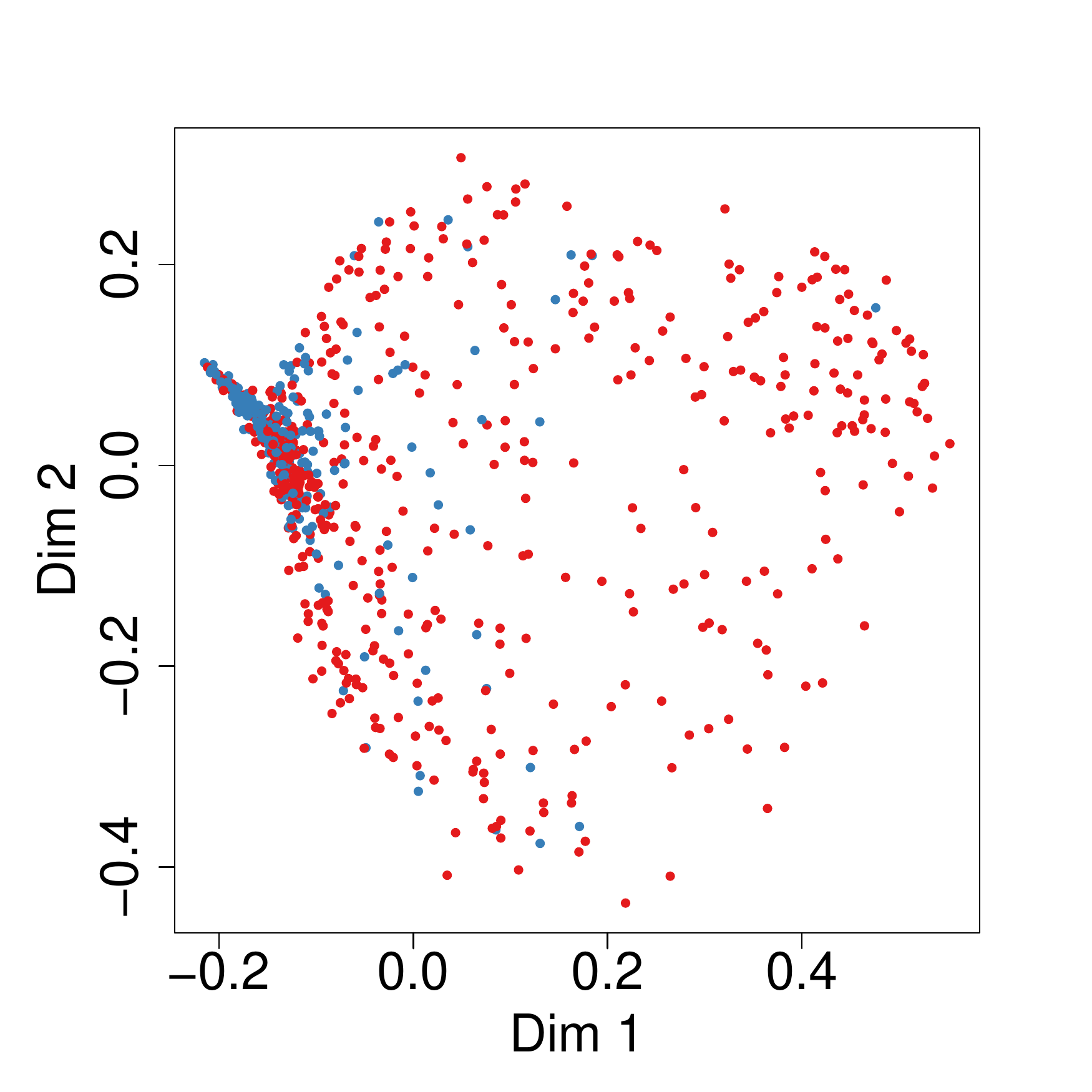}	
\end{subfigure}
\caption{The visualizations of 'Pima' data set with (a) the SOM, (b) the RF-SOM, (c) the MDS with Euclidean distance, (d) the RF-MDS. The class information is coded in red color for the first class and blue color for the second class in all plots. In the (a) and (b) for polar area diagram the attribute's number corresponding to the neuron's weight is shown in the legend.} \label{mds_som}
\end{figure}

To present visulization properties of the proposed method we used 'Pima' data set. In the Fig.\ref{mds_som} there are presented: the SOM (Fig.\ref{som_pi}) and the MDS (Fig.\ref{mds_pi}) both learned with Euclidean distance; and RF-SOM (Fig.\ref{rfsom_pi}) and RF-MDS (Fig.\ref{rfmds_pi}) constructed using the RF proximity matrix. The SOM networks were presented as a 2D grid of neurons, where for each neuron, its weigths are presented by a polar area diagram (sometimes called coxcomb plot). In the MDS and the RF-MDS plots information about points distribution in reduced 2D space is available. However, information about how point's position is affected by combination of attributes values is missing. What is more, there is hard to find crisp border to distinguish two classes on the MDS neither on the RF-MDS. In contrary to MDS technique, the SOM and RF-SOM plots do not provide information about explicit point distribution but rather a mapping of attributes combination onto 2D grid of neurons. After network labeling the class labels are assigned to neurons, therefore distinguishing specific combination of attributes in each class is possible. As expected, although, the SOM and the RF-SOM started learning from the same initial weights values they have different final distribution of attributes combinations across the network.

To compare the accuracy of the SOM learned with Euclidean distance and RF-SOM learned with RF proximity matrix we use information about samples class label. We measure the accuracy of classification. The results of comparison are presented in the Table \ref{results}. All of the results are mean over 10-fold cross validation. In the Table \ref{results} we also include the accuracy of the alone RF as the reference. The RF-SOM on 4 data sets ('Glass', 'Sonar', 'Ionosphere', 'Pima') obtained better results than the SOM. The greatest improvement over the SOM was achived on 'Sonar' set, it was $12.57\%$. The same performance of the SOM and the RF-SOM was on 'Wine' and 'Iris', when on the latter the RF-SOM has higher standard deviation value. It is worth noting, that the improvement of the RF-SOM over the SOM depends on the accuracy of the RF. On data sets (like 'Sonar' or 'Glass') where the accuracy difference between the RF and the SOM is high, the RF-SOM noted large improvement over the SOM. Whereas, on data sets ('Wine' and 'Iris') with small accuracy difference between the RF and the SOM, the RF-SOM obtained the same mean accuracy as the SOM.

\begin{table}
\begin{center}
\begin{tabular}{c|c|c|c|c|c|c|c|}
\cline{2-7}  & Glass & Wine & Iris & Sonar & Ionosphere & Pima \\ \cline{1-7}
\multicolumn{1}{|c|}{RF} & 77.96$\pm$7.82 & 98.85$\pm$2.29 & 95.33$\pm$6.70 & 85.05$\pm$4.74 & 93.44$\pm$2.88 & 76.14$\pm$7.09 \\ \hline
\multicolumn{1}{|c|}{SOM}   & 61.73$\pm$6.18 & 96.60$\pm$3.73 & 94.67$\pm$4.00 & 67.69$\pm$8.97 & 84.33$\pm$6.43 & 71.73$\pm$5.16
 \\ \hline
\multicolumn{1}{|c|}{RF-SOM} & 67.27$\pm$6.04 & 96.60$\pm$3.73 & 94.67$\pm$5.81 & 80.26$\pm$5.17 & 89.72$\pm$6.42 & 74.71$\pm$6.80 \\ \hline
\end{tabular}
\end{center}
\caption{The classification accuracy for RF, SOM and RF-SOM. The results are mean and std. over 10-fold cross validation.}\label{results}
\end{table}

\begin{figure}
\centering
\begin{subfigure}[b]{0.49\textwidth}	
	\includegraphics[clip=true, width=\textwidth]{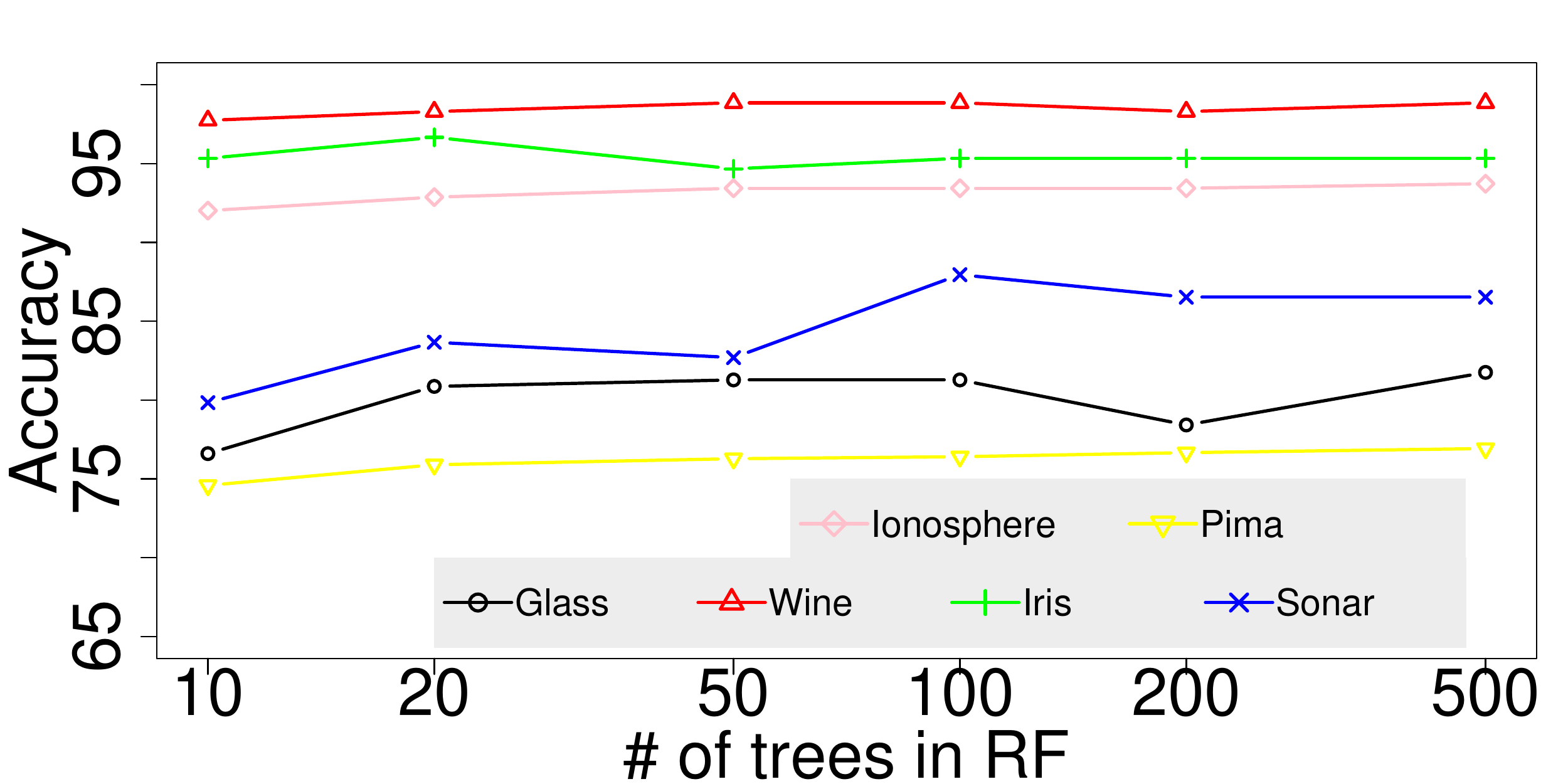} 
	\caption{RF} \label{rf_rf}	
\end{subfigure}
\begin{subfigure}[b]{0.49\textwidth}	
	\includegraphics[clip=true, width=\textwidth]{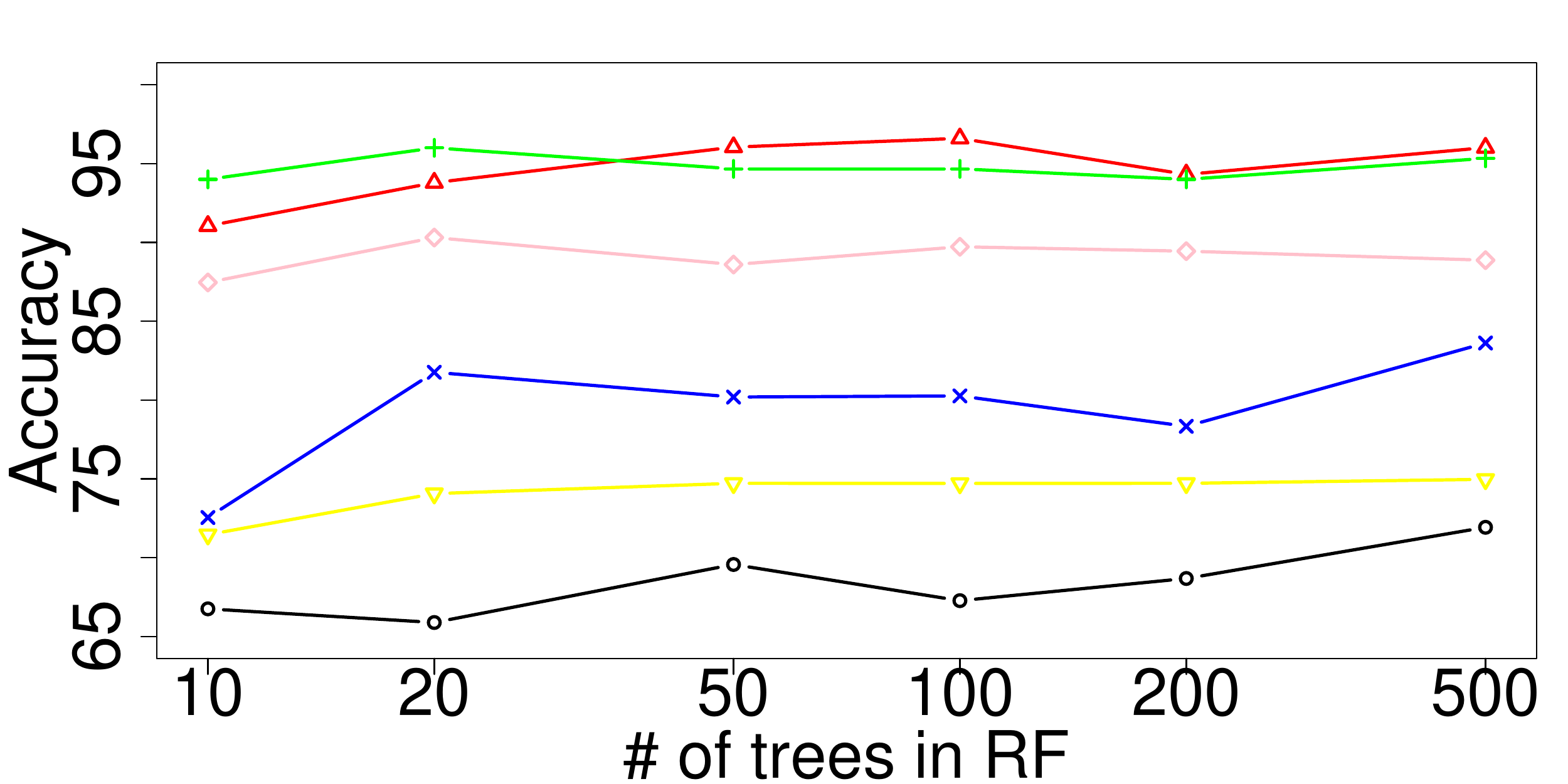} 
	\caption{RF-SOM} \label{rf_som}
\end{subfigure}
\caption{The classification accuracy for (a) RF and (b) RF-SOM, for a different number of trees in the RF. The result are mean over 10-fold cross validation.} \label{diff_rf_som}
\end{figure}

We measure classfication accuracy for different number of trees in the RF to examine the influence of a number of trees in the RF to performance of the RF-SOM. The accuracy for the RF and the RF-SOM for a number of trees in the RF, $T=\{10, 20, 50, 100, 200, 500\}$, is presented in Fig.\ref{diff_rf_som}. It can be observed that for all data sets except 'Sonar' and 'Glass' the accuracy of the RF does not depend on $T$. The good results of classification are obtained even for 10 trees in the RF. For 'Sonar' and 'Glass' sets the accuracy of the RF increases with increasing a number of trees. The very similar behaviour can be observed for the RF-SOM. The accuracy for 'Wine', 'Iris', 'Ionosphere' and 'Pima' slightly varies with $T$ growing. However, for 'Sonar' and 'Glass' sets the RF-SOM obtains better results if more trees are used in the RF. The observed behaviour can be explained by more complex data relationships in 'Sonar' and 'Glass' sets which are better modeled with greater number of trees in the RF. 

\section{Conclusions}

The novel method for visualizing the RF proximity matrix by the SOM was proposed. The RF-SOM method uses the RF to compute distances between input sample and neurons. The proposed method of visualization provide better understanding of relationship between data in the RF structure than the MDS. The RF-SOM contrary to the MDS provides a mapping of data onto 2D neurons grid. In case of new coming samples there is no need to recompute the whole RF proximity matrix like in the MDS method. Additionally, the proposed method has lower memory complexity than the MDS, which for large data sets is not applicable. What is more, the experimental results show that the RF-SOM learned with the RF dissimilarity gained better or the same accuracy than the SOM learned with Euclidean distance. As pointed in \cite{horvath} the RF dissimilarity has attractive features: it can handle mixed variable types well, is invariant to monotonic transformations of the input variables and is robust to outliers. The attractiveness of RF dissimilarity and obtained results with RF-SOM encourage to focus our future work on using the RF proximity matrix in other clustering algorithms. 

\section{Acknowledgements}

PP has been supported by the European Union in the framework of European Social Fund through the Warsaw University of Technology Development Programme.

\end{document}